\pdfoutput=1
\documentclass[11pt]{article}
\usepackage[]{EMNLP2023}

\usepackage{times}
\usepackage{latexsym}
\usepackage{pgfplots}
\usepackage{booktabs}
\usepackage{multirow}
\pgfplotsset{compat=1.17}
\usepackage{pgf-pie}

\usepackage[T1]{fontenc}
\usepackage[utf8]{inputenc}
\usepackage{graphicx}

\usepackage{microtype}
\usepackage{xspace}

\usepackage{inconsolata}
\usepackage{booktabs}
\usepackage{fdsymbol}

\newcommand{\dataset}{\texttt{TheoremQA}\xspace}

\title{TheoremQA: A Theorem-driven Question Answering Dataset}

\author{$^{\spadesuit}$Wenhu Chen\thanks{\quad Authors ordered by contribution. Corresponding author email: wenhuchen@uwaterloo.ca}, $^{\varheartsuit}$Ming Yin, $^{\spadesuit}$Max Ku, $^{\vardiamondsuit}$Pan Lu, $^{\vardiamondsuit}$Yixin Wan, \\
\textbf{$^{\spadesuit}$Xueguang Ma, $^{\varheartsuit}$Jianyu Xu, $^{\varheartsuit}$Xinyi Wang, $^{\vardiamondsuit}$Tony Xia}\\
University of Waterloo, Canada$^{\spadesuit}$ \\
University of California, Santa Barbara, United States$^{\varheartsuit}$ \\
University of California, Los Angeles, United States$^{\vardiamondsuit}$ \\}

\begin{document}
\maketitle
\begin{abstract}
The recent LLMs like GPT-4 and PaLM-2 have made tremendous progress in solving fundamental math problems like GSM8K by achieving over 90\% accuracy. However, their capabilities to solve more challenging math problems which require domain-specific knowledge (i.e. theorem) have yet to be investigated. In this paper, we introduce TheoremQA, the first theorem-driven question-answering dataset designed to evaluate AI models' capabilities to apply theorems to solve challenging science problems. \dataset is curated by domain experts containing 800 high-quality questions covering 350 theorems\footnote{e.g. Taylor's theorem, Lagrange's theorem, Huffman coding, Quantum Theorem, Elasticity Theorem, etc} from Math, Physics, EE\&CS, and Finance. We evaluate a wide spectrum of 16 large language and code models with different prompting strategies like Chain-of-Thoughts and Program-of-Thoughts. We found that GPT-4's capabilities to solve these problems are unparalleled, achieving an accuracy of 51\% with Program-of-Thoughts Prompting. All the existing open-sourced models are below 15\%, barely surpassing the random-guess baseline. Given the diversity and broad coverage of \dataset, we believe it can be used as a better benchmark to evaluate LLMs' capabilities to solve challenging science problems. 
%The data and code are released at https://github.com/wenhuchen/TheoremQA.
\end{abstract}

\section{Introduction}
A long-standing goal of AI systems is to help human beings solve challenging problems, especially more domain-specific problems. To benchmark the progress towards this goal, researchers propose to evaluate AI systems' performance on different math word problem (WMP) datasets.  In recent years, there has been a plethora of WMP datasets~\cite{lu2023dl4math}, which we include in~\autoref{tab:math_dataset}. Most of these datasets are meant for fundamental questions aimed at Grade 1-12 students on a narrow subject. On the other hand, these datasets do not involve much domain-specific knowledge, aka \textbf{theorem}. Due to these two deficiencies, we believe that these datasets are not ideal to benchmark the existing powerful LLMs~\cite{brown2020language,tamkin2022task,chen2021evaluating,chowdhery2022palm,hoffmann2022training,taylor2022galactica} due to their simplicity. In fact, on the popular GSM8K dataset~\cite{cobbe2021training}, GPT-4~\cite{gpt4} and PaLM-2~\cite{palm2} both already achieved 92\% accuracy. Similarly, we tested GPT-4~\cite{gpt4} on the subsets of several other listed datasets in~\autoref{tab:math_dataset} and observed 90+\% accuracy in most cases. The only exception is MATH~\cite{hendrycks2measuring} containing high-school math competition problems with SoTA performance around 50\%~\cite{zheng2023progressive}. However, MATH~\cite{hendrycks2measuring} is focused on math skills rather than theorem.
\begin{figure*}
    \centering
    \includegraphics[width=0.9\linewidth]{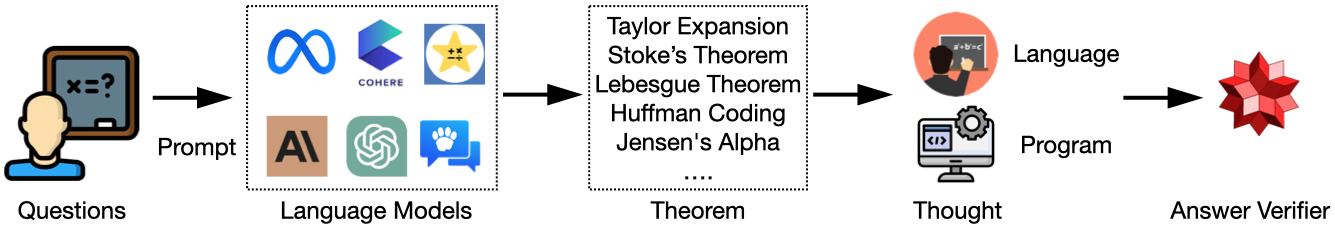}
    \caption{The overview of \dataset and the prompting strategies adopted.}
    \label{fig:overview}
\end{figure*}

\begin{table*}[!t]
    \small
    \centering
    \begin{tabular}{lcccc}
    \toprule
        Dataset                                        & Domain                  & Level                &  Source          &  Theorem \\
    \midrule
        DRAW~\cite{upadhyay2015draw}                   & Algebra                 & Elementary School    &  Generated       &   -        \\
        MAWPS~\cite{koncel2016mawps}                   & Arithmetic              & Elementary School    &  Generated       &   -        \\
        DRAW1K~\cite{upadhyay-chang-2017-annotating}   & Algebra                 & Elementary School    &  Generated       &   -        \\
        ASDiv~\cite{miao2020diverse}                   & Arithm/Algebra          & Elementary School    &  Internet        &   -        \\
        SVAMP~\cite{patel2021nlp}                      & Arithm/Algebra          & Elementary School    &  ASDiv           &   -        \\
        Math23K~\cite{wang2017deep}                    & Algebra                 & Elementary School    &  Internet        &   -        \\
    \midrule
        TabMWP~\cite{lu2023dynamic}                    & Arithm/Algebra          & Elem./Middle School  &  Textbooks       &   NO      \\
        GSM8K~\cite{cobbe2021training}                 & Arithm/Algebra          & Middle School        &  Annotated       &   NO      \\
        GEOS~\cite{seo-etal-2015-solving}              & Geometry                & Middle School        &  SAT             &   NO      \\
        Geometry3K~\cite{lu-etal-2021-inter}           & Geometry                & Middle/High School   &  Textbooks       &   NO      \\
        GeoQA~\cite{chen2021geoqa}                     & Geometry                & Middle/High School   &  Exam            &   NO      \\
        UniGeo~\cite{chen-etal-2022-unigeo}            & Geometry                & Middle/High School   &  Textbooks       &   NO      \\
        ScienceQA~\cite{lu2022learn}                   & Science                 & Middle/High School   &  Textbooks       &   NO      \\
        MATH~\cite{hendrycks2measuring}                & Math                    & High School          &  Competition     &   YES     \\
    \midrule
        AQuA~\cite{ling2017program}                    & Arithm/Algebra          & University           &  GMAT/GRE        &   NO       \\
        MathQA~\cite{amini2019mathqa}                  & Arithm/Algebra          & University           &  AQuA            &   NO       \\
        MathQA-Python~\cite{austin2021program}         & Arithm/Algebra          & University           &  AQuA            &   NO       \\
        FinQA~\cite{chen2021finqa}                     & Finance                 & University           &  CrowdSource     &   NO       \\
        TAT-QA~\cite{zhu2021tat}                       & Finance                 & University           &  CrowdSource     &   NO       \\
    \midrule
        \dataset  (Ours)                               & STEM                    & University           &  Internet+Expert &   350+      \\
    \bottomrule
    \end{tabular}
    \caption{List of existing Math and STEM QA datasets.}
    \label{tab:math_dataset}
\end{table*}

In this paper, we propose the first theorem-driven QA dataset built on university-level theorems across Math, Physics, EE\&CS, and Finance. The whole collection process takes two steps: (1) we first enumerate roughly 400 theorems in different subfields like algebra, number theory, graph theory, information theory, etc, (2) we ask domain experts to search for questions regarding these theorems from different sources like Internet and Textbooks. The domain experts will adjust these questions to ensure the answers follow the desired format for the ease of automatic evaluation. Through the careful construction process, we collected 800 high-quality question-theorem-answer triples as our final release version. 

We evaluate a wide spectrum of instruction-finetuned language and code models including GPT~\cite{brown2020language}, Claude~\cite{bai2022constitutional}, LLaMA~\cite{touvron2023llama}, Pythia~\cite{biderman2023pythia}, CodeGen~\cite{nijkamp2022codegen}, GLM~\cite{zeng2022glm}, StarCoder~\cite{li2023starcoder}, and CodeT5+~\cite{wang2023codet5+} on our dataset. We adopt two prompting methods: Chain-of-Thoughts (CoT)~\cite{weichain} and Program-of-Thoughs (PoT)~\cite{chen2022program} to prompt the large language models. We also investigate how to infuse the theorem into the thought process of LLMs and how to present the multimodal inputs to the LLMs. 

In the course of our experiments, several notable observations were made. First, GPT-4~\cite{gpt4} significantly outperformed all existing models, reaching an accuracy level of 51\% when combined with Program-of-Thoughts prompting. Trailing behind GPT-4, the second most effective model was ChatGPT, achieving an accuracy of 35\% through the same prompting method. Additionally, our human evaluation determined that half of GPT-4's errors are caused by minor mistakes like calculation errors, rounding errors, etc. We believe these errors could be easily rectified with a more deliberate prompting strategy or human intervention. This suggests that there is still significant headroom for GPT-4 to achieve with more deliberate prompting strategies. Secondly, we found that all open-source, instruction-tuned language and code models scored below 15\% in accuracy, barely exceeding the random guess baseline of 10\%. Our human evaluation reveals that open-source models like Alpaca are making errors mainly due to their ignorance of the theorem, where 90\% of the errors are not rectifiable. This stark gap between GPT and open-source models suggests that further enhancement strategies, such as science-focused pre-training or fine-tuning, should be considered to narrow the performance disparity. Thirdly, we explored the potential to do theorem-augmented generation. However, the simple strategy of concatenation did not yield a significant improvement. We conjecture that a more complex integration strategy may be needed to achieve more gains. Lastly, we examined the performance of various multi-modal instruction-tuned models on the multimodal subset of the \dataset dataset. Surprisingly, these models did not demonstrate significant performance gains over their text-only counterparts. This is mainly due to the unnaturalness of the image, which consists of lots of diagrams and text. Such images are not well captured by existing visual encoder models.

To sum up, our contributions are three folds: 
\begin{itemize}
    \item We propose the first theorem-driven question-answering dataset to understand LLMs' capabilities to apply science theorems.
    \item We comprehensively evaluate a wide spectrum of 16 LLMs on \dataset.
    \item We perform different analyses in the theorem integration and multimodal understanding aspects to provide detailed insights.
\end{itemize}

\section{Related Work}
\subsection{Math Word Problems} 
Mathematical reasoning skills are crucial for general-purpose intelligent systems, garnering significant interest from the research community. In the past, studies have explored the ability of NLP models to solve arithmetic and algebraic problems~\cite{hosseini2014learning,koncel2015parsing,roy2015solving,ling2017program}. More recently, researchers have introduced increasingly challenging datasets~\cite{saxton2019analysing,miao2020diverse,amini2019mathqa,hendrycks2measuring,lu2021iconqa,patel-etal-2021-nlp} aimed at enhancing difficulty, diversity, and adversarial robustness. LiLA~\cite{Mishra2022Lila} proposes to assemble a vast collection of mathematical datasets into a single, unified dataset. LiLA also annotates Python programs as target outputs for solving mathematical problems. However, the existing datasets were mostly focused on grade school simple mathematics. To further investigate the LLMs' capabilities to assist humans to solve challenging math problems, we propose \dataset as the first benchmark to enable research in this direction. 

\subsection{Large Language Models} 
In recent years, there has been a surge of research and development in the area of large language models (LLMs) that have significantly advanced the field of natural language processing. GPT-3~\cite{brown2020language} demonstrated a strong capability to perform few-shot predictions, where the model is given a description of the task in natural language with few examples. By using human-feedback reinforcement learning, InstructGPT~\cite{ouyang2022training} has shown its unprecedented capabilities to follow human instructions. Scaling model size, data, and computing are crucial to enable this learning ability. Later, ~\citet{rae2021scaling, chowdhery2022palm, zhang2022opt,touvron2023llama,chen2021evaluating} have proposed to train different types of LLMs with different training recipes. The capability to follow few-shot exemplars to solve unseen tasks is not existent on smaller LMs, but only emerges as the model scales up~\cite{weiemergent}. More recently, GPT-4~\cite{gpt4} shows tremendous progress on lots of complex reasoning tasks spanning mathematics, coding, vision, medicine, law, psychology, and more. \citet{bubeck2023sparks} shows that GPT-4 is already demonstrating more general intelligence than previous AI models. To further validate GPT-4's capability to solve challenging reasoning tasks, we propose \dataset as the new benchmark to further understand LLMs' upper limit.

\subsection{Reasoning with Large Language Model}
To better unleash large language models' capabilities to solve complex reasoning tasks. Chain-of-Thought Prompting~\cite{weichain,kojimalarge,wang2022self} was proposed, which aims at prompting the large language models to generate the `thought process' before outputting the answer. Later on, several other works~\cite{drozdov2022compositional,zhou2022least,nyeshow} also propose different approaches to utilize LLMs to solve reasoning tasks by allowing intermediate steps. Our method can be seen as an extension to CoT by leveraging an extra step of symbolic execution. Another line of work~\cite{gao2022pal,chen2022program} was proposed to adopt Python programs as the demonstration for the `thought process' to solve different reasoning tasks.

\section{Dataset}
\begin{figure*}
    \centering
    \includegraphics[width=0.82\linewidth]{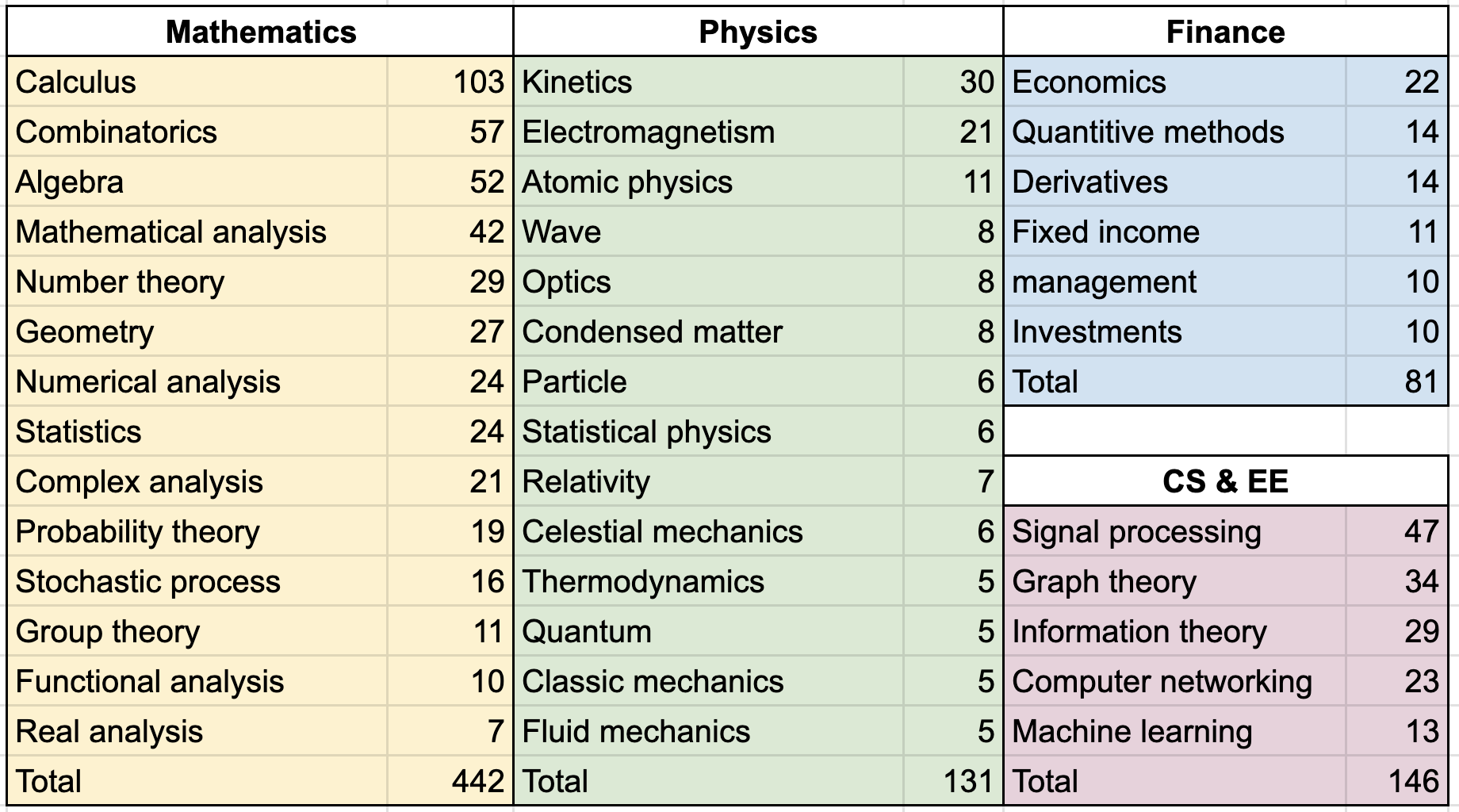}
    \caption{Subfields of \dataset under Math, Physics, Engineering, and Finance.}
    \label{fig:subfield}
\end{figure*}

Our dataset collection pipeline contains two steps:

\paragraph{Theorem Enumeration}
Our aim was to encompass a wide range of theorems. To this end, we began by prompting Large Language Models (LLMs), specifically GPT-4~\cite{gpt4}, to enumerate popular subfields in Mathematics, Physics, Finance, and Electrical Engineering \& Computer Science. The covered subfields are listed in~\autoref{fig:subfield}. Subsequently, we prompted GPT-4 to propose plausible university-level theorems relevant to these subfields. For instance, within the 'Calculus' subfield, GPT-4 might suggest the 'Intermediate Value Theorem', 'Rolle's Theorem', and so on. After gathering an extensive list of theorems, we assembled a team of domain experts (holders of Masters and PhDs in Statistics, Electrical Engineering, Computer Science, and Finance) to refine the theorem inventory and supplement any omitted theorems. Ultimately, we collected approximately 400 theorems, encapsulating a diverse range of topics within these fields. We then delegated these theorems to nine domain experts, instructing them to locate question/answer pairs from varied sources. During the annotation process, a small number of theorems were discarded due to their evaluation complexity.

\begin{figure}[!t]
    \centering
    \includegraphics[width=0.85\linewidth]{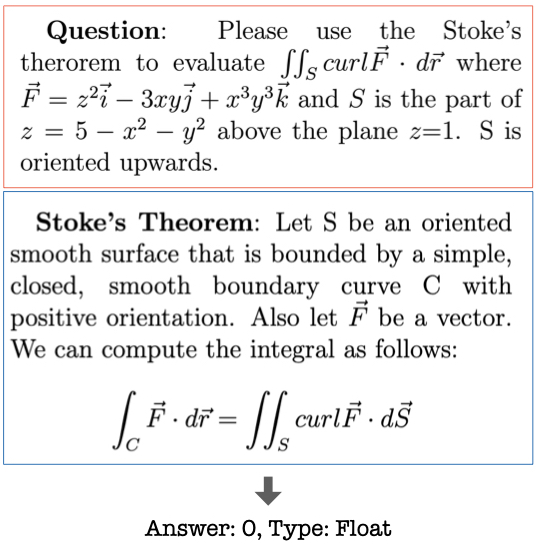}
    %\vspace{3ex}
    \includegraphics[width=0.85\linewidth]{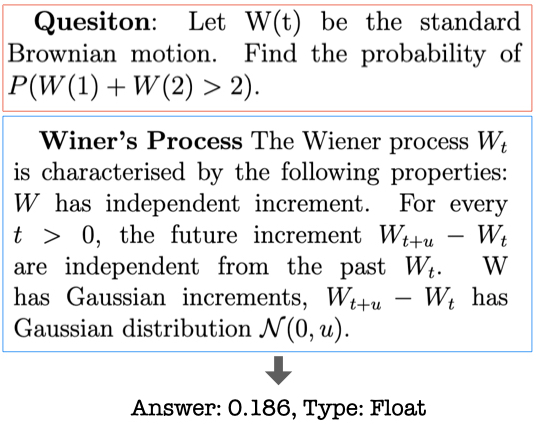}
    \caption{Examples from \dataset. The first question requires the usage of Stoke's theorem to transform the double integral into a line integral. The second question requires knowing the properties of Wiener's process.}
    \label{fig:example}    
\end{figure}

\paragraph{Question Annotation}
Our problems were sourced from websites, books, or devised by the experts themselves. One challenge we encountered was the potential for questions found online to have been included in the training data. To mitigate this 'data contamination' issue, we encouraged domain experts to modify these questions. Another challenge arose from questions with answers in symbolic form, matrix form, figure form, etc. These presented significant obstacles for our automatic evaluation. To overcome this, we instructed domain experts to alter the question so the answer would be limited to the following forms: (1) integer, (2) float, (3) list of integers/floats, (4) boolean, and (5) multiple-choice options. For instance, if the original question concerned a matrix, we would revise it to ask about the trace of the answer matrix. This modification significantly streamlined the evaluation process. An example of this can be found in~\autoref{fig:example}.

%\paragraph{Answer Verification}
%Finally, we prompt GPT-4 to generate answers to these questions and identify the ones which are highly different from the annotated answers. We suspect that these questions might contain certain ambiguities or errors. We do another pass over these suspicious questions to involve further correction and adjustment. We show the subfields of these theorems in~\autoref{fig:subfield}. 

\begin{figure}
    \centering
    \begin{tikzpicture}
[scale=0.7]
\pie{
47/float,
27/integer,
15/bool,
9/list,
2/option
}
\end{tikzpicture}
    \vspace{-6ex}
    \caption{Answer type distribution in \dataset.}
    \label{fig:answer_distribution}
\end{figure}

\paragraph{Dataset Statistics}
Finally, we collected a total of 800 questions over 354 theorems. Specifically, there are 199 Math theorems, 52 Physics theorems, 55 Finance theorems, and 48 CS\&EE theorems. There are 442 Math questions, 146 CS\&EE questions, 131 physics questions, and 81 Finance questions. We show the answer-type distribution in~\autoref{fig:answer_distribution}. To further enhance the multimodality aspect of \dataset, we also include 51 questions with image input (diagrams), where the model needs to understand the visual input to answer questions. 

The majority of the questions in \dataset have float and integer as the answers, which is more realistic than the existing multi-choice datasets like ScienceQA~\cite{lu2022learn} or AQuA QA~\cite{ling2017program}. Therefore, the models are unlikely to take shortcuts to achieve high accuracy. 

\paragraph{Human-Level Performance}
To provide a rough but informative estimate of human-level performance. We randomly select 20 questions and assign these questions to the 4 Math\&CS undergraduate students (average GPA) who have taken the required courses regarding these questions. The participants are given 24 hours with internet access to solve these questions. The four undergraduate students achieve 12/20, 15/20, 18/20, and 19/20 scores on these randomly sampled questions. From this experiment, we are more confident that an expert-level performance should be 100\%.

\section{Method}
Our method for addressing these demanding questions in the \dataset is comprised of several distinct modules, as outlined in~\autoref{fig:overview}:

\paragraph{Prompting}
We utilize two established prompting strategies:
\begin{itemize}
\item Chain-of-Thought Prompting~\cite{weichain}: This strategy prompts the language model to initially generate a step-by-step thought process, eventually leading to the final answer.
\item Program-of-Thought Prompting~\cite{chen2022program,gao2022pal}: This strategy prompts the language model to progressively generate a program. The final answer is then derived by executing this program.
\end{itemize}
By delegating computational tasks to an external executor, the problem-solving process is considerably enhanced in its reliability. This improvement results in remarkable advancements in existing math datasets being reported in~\cite{chen2022program}.

\paragraph{Answer Extraction}
We observed that parsing the output from Large Language Models (LLMs) can be challenging due to two main issues: (1) The answer is often embedded within a sentence, making it difficult to extract using regular expressions, and (2) The answer may not be normalized, such as 'pi / 3' or '2*10 - e', which complicates comparison with the ground truth. To tackle these problems, we initially employ ChatGPT to identify the answer span within the model's output, then forward this string to WolframAlpha~\cite{Mathematica} for normalization into a float, integer, or list.

\paragraph{Theorem Augmentation}
We explored the potential of enhancing large language models with retrieved theorem descriptions to assess their effect on performance. One approach is to retrieve descriptions of the given theorems from the Internet to supplement the LLMs' output. Another experiment involved prompting GPT-4 to generate text descriptions of the theorem, which are then used as an additional augmentation signal.

\paragraph{Multimodal Input}
A small portion of our data (50 instances) includes images, such as diagrams, as supplemental input, particularly in geometry questions. Since current LLMs don't support such multimodal inputs, we propose a solution: to employ captions like Chameleon~\cite{lu2023chameleon}. These captions describe the image and are then appended to the LLMs' output as an additional signal.

\section{Experiments}
\subsection{Model Descriptions}
In our experiments, we mainly investigate the following models:
\begin{itemize}
    \item GPT3/3.5/ChatGPT/GPT4: These are instruction-tuned models from OpenAI\footnote{https://openai.com/}. 
    \item PaLM-2: This is the instruction-tuned model from Google~\cite{palm2}.
    \item Claude-v1/Claude-instant: These are instruction-tuned models from AnthropicAI\footnote{https://www.anthropic.com/index/introducing-claude}.
    \item Alpaca-13B: This model is based on the LLaMA~\cite{touvron2023llama}. Alapaca is instruction-tuned by the 52K data generated from GPT-4.
    \item Vicuna-13B: This model is based on the LLaMA~\cite{touvron2023llama}. Vicuna is instruction-tuned by the 100K ShareGPT data generated by different GPT-based models.
    \item OpenAssistant-12B: This model is based on Pythia~\cite{biderman2023pythia}. The model is instruction-tuned by OpenAssistant data\footnote{https://open-assistant.io/}.
    \item MOSS-instruct-16B: This model is based on CodeGen~\cite{nijkamp2022codegen}, which is further instruction-tuned with instruction following dataset distilled from GPT.\footnote{https://txsun1997.github.io/blogs/moss.html}. 
    \item StarChat-16B: This model is based on StarCoder~\cite{li2023starcoder}. StartChat is being instruction-tuned on OpenAssistant data\footnote{https://open-assistant.io/} and ShareGPT data.
    \item InstructCodeT5+: This model is based on CodeT5+~\cite{wang2023codet5+}. InstructCodeT5+ is further instruction-tuned on Code Alpaca data\footnote{https://github.com/sahil280114/codealpaca} to follow instructions.
\end{itemize}

\subsection{Main Results}
We demonstrate our main results on~\autoref{tab:results}. We will summarize different findings in the following:
\paragraph{Closed-source Models}
For GPT-3 (text-davinci-002) and GPT-3.5 model, since these two models are not Chat-based models, we need to demonstrate one example ensure to help them generate outputs of the desired format. With CoT prompting, GPT-3 (text-davinci-002) and GPT-3.5 models are only achieving 16.6\% and 22.8\% accuracy. By adopting the program as the intermediate reasoning form, both models can gain reasonable improvements. For Claude-v1, we found that it is matching the performance of GPT-3.5. ChatGPT outperforms GPT-3.5 and Claude-v1 significantly by 8\%, which indicates ChatGPT's capabilities to perform complex numerical reasoning. GPT-4 is the strongest model being evaluated, which beats all the rest models by a huge margin. With Chain-of-Thoughts prompting, GPT-4 can outperform ChatGPT by 13\%. With Program-of-Thoughts prompting, GPT-4 can outperform ChatGPT by 16\%. Though some other models have shown to match GPT-4 on simple tasks, GPT-4's capability to solve challenging tasks seems unparalleled. 

\paragraph{Open-source Models}
For the open-source models, we found that their performance is much behind. To better understand their accuracy, we also provide the random-guess baseline of 10\%. We test both prompting strategies, however, their results consistently lie in the range of 10-14\%. The results indicate that these open-source LMs are still struggling with more complex mathematical reasoning tasks in \dataset. Given that ChatGPT of a similar size is able to achieve much higher performance, we believe the parameter size is not the only cause. There is still a significant amount of effort during pre-training or supervised fine-tuning to instill enough scientific knowledge into the models' parameters to close the gap.

\paragraph{Program of Thoughts Analysis}
From~\autoref{tab:results}, we observe that PoT brings consistent improvement over CoT on GPT-* models. Different GPT-* models can normally yield a gain of 5-8\% accuracy. In contrast, Claude-v1 and StarChat are almost obtaining the same accuracy. To better analyze where the gains are coming from, we plot~\autoref{fig:execution_acc} to understand how many of generated Python programs are actually `executable'. As can be seen, both StarChat and CodeT5+ are having trouble generating `runnable' programs with only 40\% programs being executable. Claude-v1 is able to increase the validity of the generated programs to 60\%. In contrast, GPT3.5 and ChatGPT can further increase the ratio to around 80\%. GPT-4 is extremely accurate in generating programs, where 92\% of the generated programs are runnable. Such a high executable ratio explains why the gain brought to GPT-* model is much higher than Claude-v1 and StarChat.

%From~\autoref{tab:math_dataset}, we can see that Physics is actually the hardest subject. The open-source models are obtaining less than 4\% accuracy on the Physics problems. There are several reasons: (1) the language models are generally lacking much physics knowledge, which leads to using the wrong theorems for computation, and (2) Physics questions involve computation over high-digit numbers, which makes the language models more prone to errors.

\begin{table*}[!t]
\centering
\small
\begin{tabular}{l|ccccc|cccc|l}
\toprule
Model               & Integer & Float & Option & List & Bool & Math & CS\&EE & Physics & Finance & All    \\
\midrule
Random Guess        & 0       &  0    & 38.9   & 0    & 65.5 & 10.0 & 24.7   & 0       &  4.9    & 10.5          \\
\midrule
\multicolumn{11}{c}{Chain of Thoughts (CoT)} \\
\midrule
GPT-3               & 11.6    & 11.7  & 27.8   & 6.8  & 46.6 & 15.8 & 34.2   & 2.3     &  12.3   & 16.6          \\
GPT-3.5             & 13.0    & 14.3  & 50.0   & 13.7 & 69.8 & 22.6 & 36.3   & 7.6     &  23.5   & 22.8          \\
ChatGPT             & 32.4    & 22.3  & 50.0   & 20.5 & 55.2 & 31.0 & 41.1   & 16.8    &  28.4   & 30.2          \\
GPT-4               & 40.3    & 36.7  & 66.7   & 37.0 & 74.6 & 43.9 & 50.6   & 30.5    &  51.4   & \textbf{43.8} \\
PaLM-2              & 26.4    & 22.8  & 61.1   & 23.3 & 71.6 & 31.0 & 47.3   & 19.8    &  27.2   & 31.8          \\
Claude-v1           & 18.1    & 19.4  & 27.8   & 15.1 & 61.2 & 21.7 & 42.5   & 13.7    &  28.4   & 24.9          \\
Cluade-instant      & 19.9    & 16.7  & 44.4   & 17.8 & 53.4 & 21.5 & 36.3   & 14.5    &  27.2   & 23.6          \\
\midrule
Alpaca (13B)        & 11.1    & 6.9   & 27.8   & 2.7  & 45.7 & 12.9  & 27.4   & 3.8     &  9.9    & 13.5       \\
Vicuna (13B)        & 8.8     & 6.9   & 16.7   & 2.7  & 45.7 & 12.2  & 24.0   & 3.1     &  12.3   & 12.9       \\
OpenAssistant (12B) & 8.3     & 5.0   & 22.2   & 1.4  & 37.9 & 10.2  & 25.0   & 0       &  4.9    & 10.7       \\
MOSS (16B)          & 8.8     & 5.4   & 24.2   & 2.4  & 44.2 & 11.3  & 28.4   & 1.6     &  8.9    & 12.2      \\
StarChat (16B)      & 7.9     & 4.9   & 22.3   & 1.9  & 44.1 & 10.7  & 23.5   & 0.6     &  6.8    & 11.6       \\
\midrule
\multicolumn{11}{c}{Program of Thoughts (PoT)} \\
\midrule
GPT-3        & 17.1    & 15.9  & 22.2   & 9.6  & 49.1 & 23.3 & 25.4   & 8.4     &  17.3   & 20.6          \\
GPT-3.5      & 23.6    & 19.9  & 50.0   & 21.9 & 61.2 & 26.7 & 41.1   & 14.5    &  30.9   & 27.8          \\
ChatGPT      & 31.0    & 35.0  & 38.9   & 21.9 & 54.3 & 35.7 & 35.6   & 26.7    &  49.4   & 35.6          \\
GPT-4        & 44.4    & 50.7  & 66.7   & 39.7 & 78.4 & 52.0 & 51.4   & 45.8    &  66.7   & \textbf{52.4} \\
Claude-v1    & 17.1    & 21.8  & 33.3   & 6.9  & 62.5 & 23.1 & 37.5   & 17.1    &  28.4   & 25.9          \\
\midrule
StarChat (16B)        & 7.7     & 6.1   & 0.0    & 3.0  & 43.5 & 13.6 & 17.6   & 5.1     &  5.1   & 11.3         \\
InstructCodeT5+ (16B) & 8.9     & 6.3   & 0.0    & 6.9  & 45.2 & 13.8 & 17.9   & 4.2     &  5.1   & 11.6         \\
\bottomrule
\end{tabular}
\caption{Results for CoT and PoT prompting on \dataset. We report the accuracy over different fine-grained question types and scientific fields.}
\label{tab:results}
\end{table*}

\begin{figure}
    \centering
    \begin{tikzpicture}
\begin{axis} [
ybar,
height=0.8in, 
bar width=0.3cm,
width=0.95\linewidth,
scale only axis,
ymin = 0.3, 
ymax = 1.0,
yticklabels=\empty,
axis x line*=bottom,
hide y axis,
tick label style={yshift=-8pt},
xticklabel style = {font=\small,yshift=0.5ex},
symbolic x coords={
GPT3,
GPT3.5,
ChatGPT,
GPT4,
Claude,
StarChat,
CodeT5+
},
legend style={
    at={(0,1.0)},
    anchor=north west,
    legend columns=-1
},
xtick=data,
yticklabels=\empty,
x tick label style={rotate=20,anchor=center},
nodes near coords,
nodes near coords align={vertical},
every node near coord/.append style={font=\tiny},
]
\addplot coordinates {
(GPT3, 0.72) (GPT3.5, 0.78) (ChatGPT, 0.82) (GPT4, 0.92) (Claude, 0.60) (StarChat, 0.40) (CodeT5+, 0.36)};
\end{axis}
\end{tikzpicture}
    \vspace{-6ex}
    \caption{Ratio of Executable Python Program of different models with PoT prompting.}
    \label{fig:execution_acc}
\end{figure}

\subsection{Additional Result}
\paragraph{Theorem Augmentation}
We also investigate whether feeding theorem as an additional text condition would help the model better solve the problem. Specifically, we ask GPT-4 to generate a paragraph to describe the theorem, which we post-processed to ensure correctness. We feed the theorem in the prompt to different language models to see the performance change and plot our findings in~\autoref{tab:augment}. For all the evaluated scenarios, we found that the improvement is limited to within 1\%. Unlike the Text or KB knowledge, theorem knowledge is more abstract and symbolic, simply concatenating the theorem definition is not enough. We believe a more sophisticated augmentation scheme is needed to truly help the model understand and apply the theorems to solve problems.
\begin{table}[!h]
\centering
\small
\begin{tabular}{l|ccc}
\toprule
Model      &  Method  &  Theorem  &  All \\
\midrule
ChatGPT    &  CoT     &  -        &  30.2 \\
ChatGPT    &  CoT     &  +        &  30.8 \\
\midrule
Claude-v1   &  CoT    &  -       &  24.9 \\
Claude-v1   &  CoT    &  +       &  25.4 \\
\midrule
ChatGPT    &  PoT     &  -        &  35.6 \\
ChatGPT    &  PoT     &  +        &  35.8 \\
\midrule
Alpaca-13B &  CoT     &  -        &  13.5 \\
Alpaca-13B &  CoT     &  +        &  14.2 \\
\bottomrule
\end{tabular}
\caption{Results for CoT and PoT prompting with additional theorem conditions.}
\label{tab:augment}
\end{table}

\begin{figure}
    \centering
    \includegraphics[width=1.0\linewidth]{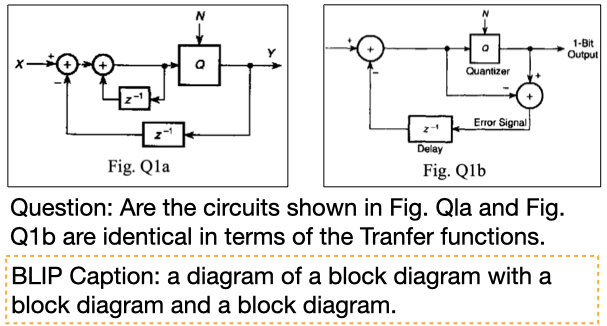}
    \caption{An example of Multimodal question.}
    \label{fig:visual_question}
\end{figure}

\paragraph{Multimodal Questions}
Our aim was to assess how effectively the current method could tackle multimodal questions (those with image inputs) in the \dataset dataset. An example is illustrated in~\autoref{fig:visual_question}, where an image is converted into 'captions' by BLIP~\cite{li2022blip}. We graphed the results from over 50 multimodal question subsets in~\autoref{fig:multimodal}. Notably, this subset posed substantial challenges; none of the models were able to achieve an accuracy rate of 10\%. This is primarily due to information loss during the captioning process.

In light of this, we conducted further evaluations on two multimodal instruction-tuned models, LLaVA-13B~\cite{liu2023visual} and VisualGLM-6B~\cite{zeng2022glm}\footnote{https://github.com/THUDM/VisualGLM-6B}. These models utilize a visual encoder (either CLIP~\cite{radford2021learning} or BLIP~\cite{li2022blip}) to encode image input, which is then integrated with language models for multimodal conversation. However, these models demonstrated performance similar to their text-only equivalent, Alpaca, with the addition of a visual encoder not significantly enhancing the results. We hypothesize that the current visual encoding modules may not be suited for representing these diagrammatic images, resulting in these less than ideal outcomes. We believe these multimodal questions remain a challenge for the research community, and we eagerly anticipate further advancements in addressing these multimodal scientific questions.

\begin{figure}
    \centering
    \begin{tikzpicture}
\begin{axis} [
ybar,
height=0.8in, 
bar width=0.3cm,
width=0.95\linewidth,
scale only axis,
ymin = 0.0, 
ymax = 15,
yticklabels=\empty,
axis x line*=bottom,
hide y axis,
tick label style={yshift=-8pt},
xticklabel style = {font=\small,yshift=0.5ex},
symbolic x coords={
Alpaca,
VisualGLM,
LLaVA,
Claude,
GPT-3,
ChatGPT,
GPT4
},
legend style={
    at={(0,1.0)},
    anchor=north west,
    legend columns=-1
},
xtick=data,
yticklabels=\empty,
x tick label style={rotate=20,anchor=center},
nodes near coords,
nodes near coords align={vertical},
every node near coord/.append style={font=\tiny},
]
\addplot coordinates {
(Alpaca, 1.8) (VisualGLM, 3.7) (LLaVA, 3.7) (Claude, 3.7) (GPT-3, 5.6) (ChatGPT, 9.3) (GPT4, 7.5)};
\end{axis}
\end{tikzpicture}
    \vspace{-6ex}
    \caption{Accuracy on the Multimodal Question Subset}
    \label{fig:multimodal}
\end{figure}

\paragraph{Error Analysis}
We conduct detailed error analysis on 200 erroneous cases from different models to analyze their error distribution. Specifically, we pick GPT4, ChatGPT and Alpaca to understand their error sources. We include the following error types: (E1) the model does not even know this theorem, (E2) the model does know the theorem, but uses the wrong formula or algorithm, (E3) the model knows the theorem and the formula, the error is only caused by minor calculation mistakes. The severity of the error decrease as the error number increases. We plot our findings in~\autoref{fig:error_analysis}, where the bar indicate the percentage of specific error types. We can observe that almost half of the errors made by GPT4 are non-critical with caused by minor calculation mistakes. This error analysis suggests that there is a still a significant headroom for GPT4 to improve with more deliberate prompting strategies or human intervention to mitigate these minor errors. In contrast, Alpaca's errors are mainly caused by not knowing the theorem at all. 

\begin{figure}
    \centering
    \begin{tikzpicture}
\begin{axis} [
ybar,
height=1.2in, 
bar width=0.45cm,
width=0.95\linewidth,
scale only axis,
ymin = 0, 
ymax = 1.2,
yticklabels=\empty,
axis x line*=bottom,
hide y axis,
enlarge x limits=0.2,
xticklabel style = {font=\small,yshift=0.2ex},
symbolic x coords={
E1,
E2,
E3
},
legend style={
    at={(0,1.0)},
    anchor=north west,
    legend columns=-1
},
%title=Breakdown Accuracy over Question Types,
xtick=data,
yticklabels=\empty,
nodes near coords,
nodes near coords align={vertical},
every node near coord/.append style={font=\tiny},
]
\addplot coordinates {
(E1, 0.278) (E2, 0.242) (E3, 0.464)
};
\addplot coordinates {
(E1, 0.355) (E2, 0.242) (E3, 0.375)
};
\addplot coordinates {
(E1, 0.75) (E2, 0.15) (E3, 0.10)
};
\legend{GPT4, ChatGPT, Alpaca}
\end{axis}
\end{tikzpicture}
    \vspace{-6ex}
    \caption{Error Analysis for GPT4, ChatGPT and Alpaca. Error severity: E1 > E2 > E3}
    \label{fig:error_analysis}
\end{figure}

\begin{figure*}[!h]
    \centering
    \includegraphics[width=0.98\linewidth]{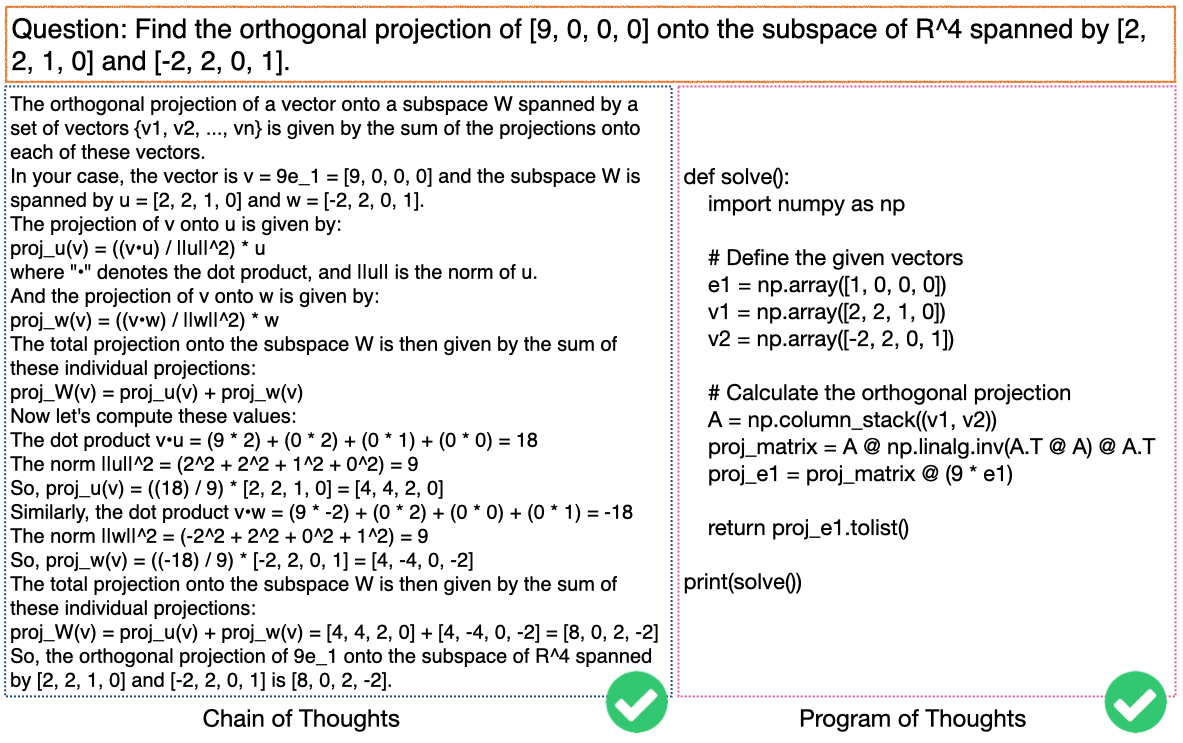}
    \includegraphics[width=0.98\linewidth]{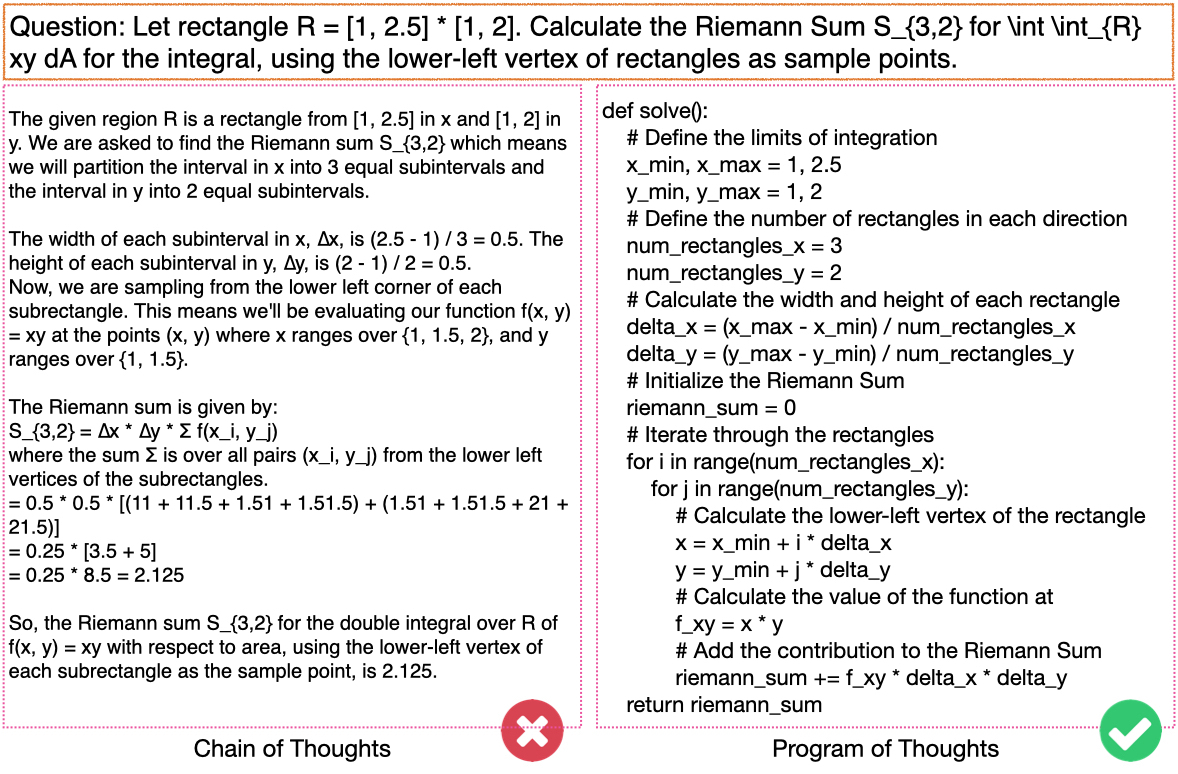}
    \caption{Case Study of GPT-4 generation with both prompting strategies.}
    \label{fig:case_study}
    \vspace{-3ex}
\end{figure*}

\paragraph{Case Study}
We list a few successful and failed examples generated by GPT-4 in~\autoref{fig:case_study} to do a side-by-side comparison between chain-of-thoughts prompting and program-of-thoughts prompting. In the first example, the question is regarding `orthogonal projection theorem'. As can be seen, Chain-of-Thoughts prompting requires a very long paragraph to generate the results. We prompted GPT-4 a few times with the same input and the results seem unstable. Sometimes the model will make tiny computation mistakes in the middle to derive the wrong answer. In contrast, the program solution is brief and concise, which leads to rather stable outputs. For the second example, the computation requires `for loop' to iteratively compute delta values for Riemann Sum. We found that such problems are also more natural for programs to solve. Through these examples, we can see GPT-4's unprecedented capabilities to solve these difficult math problems even without any demonstration or hints.   

We also show some examples in~\autoref{fig:case_study} to compare the results of CoT and PoT prompting. We can see that the PoT can significantly shorten the output sequence length. By leveraging the additional tool, PoT is able to significantly lower the task difficulty.

\section{Conclusion}
In this paper, we propose the first theorem-driven science question-answering dataset and evaluate different LLMs on it. Though GPT-4 can achieve strong performance on our new dataset, the existing open-source LLMs are still struggling to achieve reasonable performance. We conjecture it is essential to leverage more science-related pre-training or fine-tuning to close the gap. On the hand, we found that the multimodal science questions are still extremely challenging for the existing visual LLMs. We believe more specialized visual encoding models are needed to better represent diagrams in these science questions. 

\section*{Limitations}
In this work, we explore the possibilities to utilize different large language models to solve challenging theorem-driven questions. There are still some limitations: (1) our answer extraction is still not perfect. There are some cases where our answer extractor is not able to locate the answer. Thus the final accuracy is still an approximate lower bound. (2) in our dataset collection, we specifically avoid the hard-to-evaluate cases where the answer is a formula, figure, or a matrix. Our choice of the questions can be biased in terms of evaluating the overall ability. (3) in the multimodal questions in \dataset, we have investigated different existing models but none of them succeed in achieving reasonable performance. 

\bibliography{anthology,custom}
\bibliographystyle{acl_natbib}

\end{document}